\documentclass[10pt,twocolumn,letterpaper]{article}

\usepackage[pagenumbers]{cvpr} 

\usepackage{graphicx}
\usepackage{amsmath}
\usepackage{amssymb}
\usepackage{booktabs}
\usepackage{multirow}
\usepackage{balance}
\usepackage{pifont}
\usepackage{bbding}
\usepackage{soul}
\usepackage{multicol}
\usepackage{stfloats}

\usepackage[pagebackref,breaklinks,colorlinks]{hyperref}

\usepackage[capitalize]{cleveref}
\crefname{section}{Sec.}{Secs.}
\Crefname{section}{Section}{Sections}
\Crefname{table}{Table}{Tables}
\crefname{table}{Tab.}{Tabs.}

\begin{document}

\title{Long-Range Zero-Shot Generative Deep Network Quantization}

\author{Yan Luo$^1$,Yangcheng Gao$^1$, Zhao Zhang$^{1*}$, Jicong Fan$^2$, Haijun Zhang$^3$, and Mingliang Xu$^4$  \vspace{1.5mm}\\
$^1$Hefei University of Technology, China\\
$^2$The Chinese University of Hong Kong (Shenzhen), China\\
$^3$Harbin Institute of Technology (Shenzhen)\\
$^4$Zhengzhou University
}
\maketitle

\begin{abstract}
Quantization approximates a deep network model with floating-point numbers by the one with low bit width numbers, in order to accelerate inference and reduce computation. Quantizing a model without access to the original data, zero-shot quantization can be accomplished by fitting the real data distribution by data synthesis. However, zero-shot quantization achieves inferior performance compared to the post-training quantization with real data. We find it is because: 1) a normal generator is hard to obtain high diversity of synthetic data, since it lacks long-range information to allocate attention to global features; 2) the synthetic images aim to simulate the statistics of real data, which leads to weak intra-class heterogeneity and limited feature richness. To overcome these problems, we propose a novel deep network quantizer, dubbed Long-Range Zero-Shot Generative Deep Network Quantization (LRQ). Technically, we propose a long-range generator to learn long-range information intead of simple local features. In order for the synthetic data to contain more global features, long-range attention using large kernel convolution is incorporated into the generator. In addition, we also present an Adversarial Margin Add (AMA) module to force intra-class angular enlargement between feature vector and class center. As AMA increases the convergence difficulty of the loss function, which is opposite to the training objective of the original loss function, it forms an adversarial process. Furthermore, in order to transfer knowledge from the full-precision network, we also utilize a decoupled knowledge distillation. Extensive experiments demonstrate that LRQ obtains better performance than other competitors.
\end{abstract}

\begin{figure}[t]
  \centering
   \includegraphics[width=\linewidth]{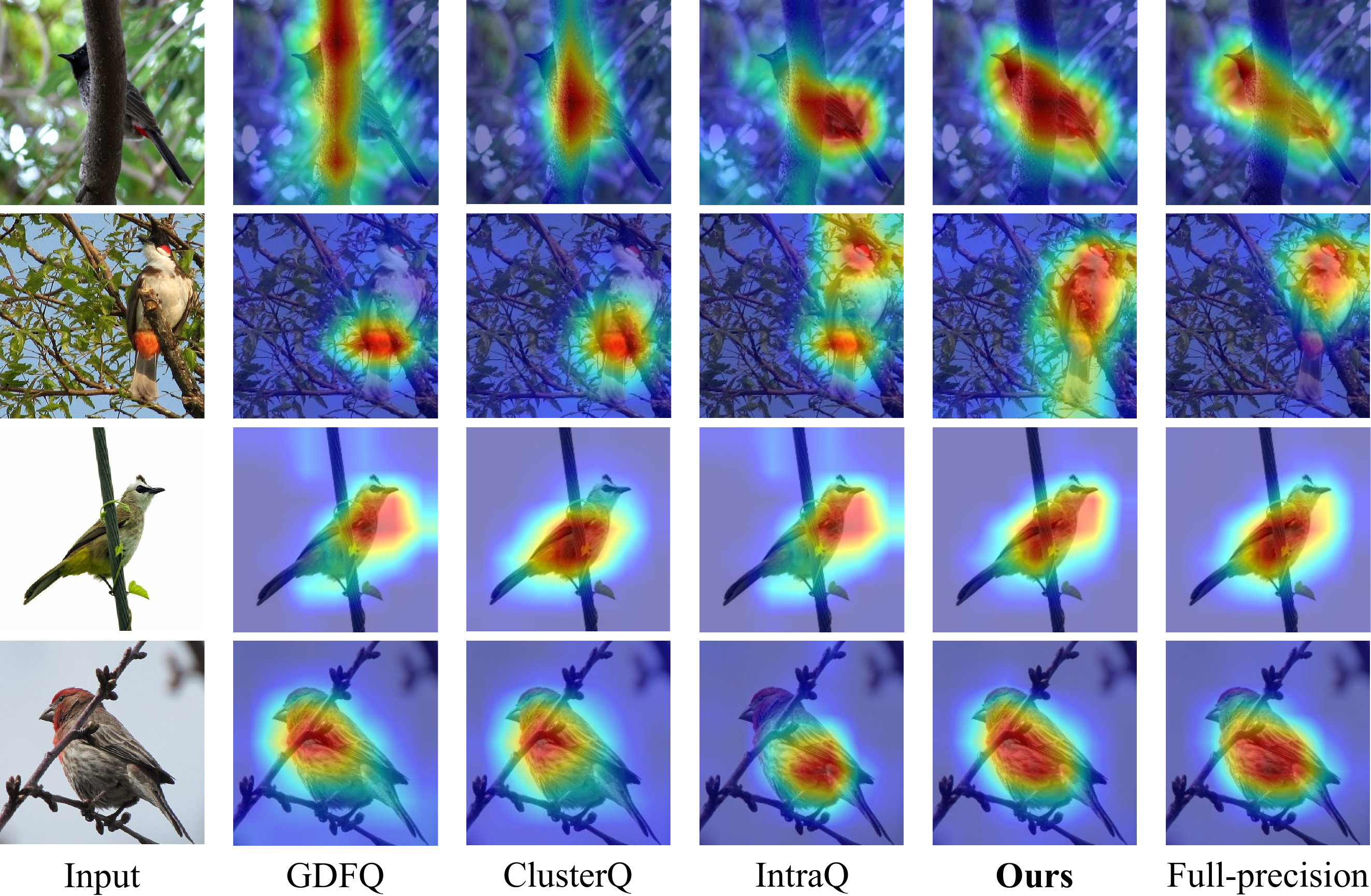}
   \caption{Visualization of the Grad-CAM \cite{selvaraju2017grad} results of the ResNet18 \cite{he2016deep} quantized to 4 bit-width by different ZSQ methods, including GDFQ \cite{xu2020generative}, ClusterQ \cite{gao2022clusterq}, IntraQ \cite{zhong2022intraq} and our LRQ. Clearly, compared to others, our LRQ obtains the best results, performing the most similar attention as full-precision model. Specifically, even if some parts of the object are occluded, the quantized model by our LRQ can still pay attention to the whole object more accurately, instead of being scattered.}
   \label{fig:attention}
   \vspace{-3mm}
\end{figure}

\begin{figure*}[t]
  \centering
   \includegraphics[width=0.96\linewidth]{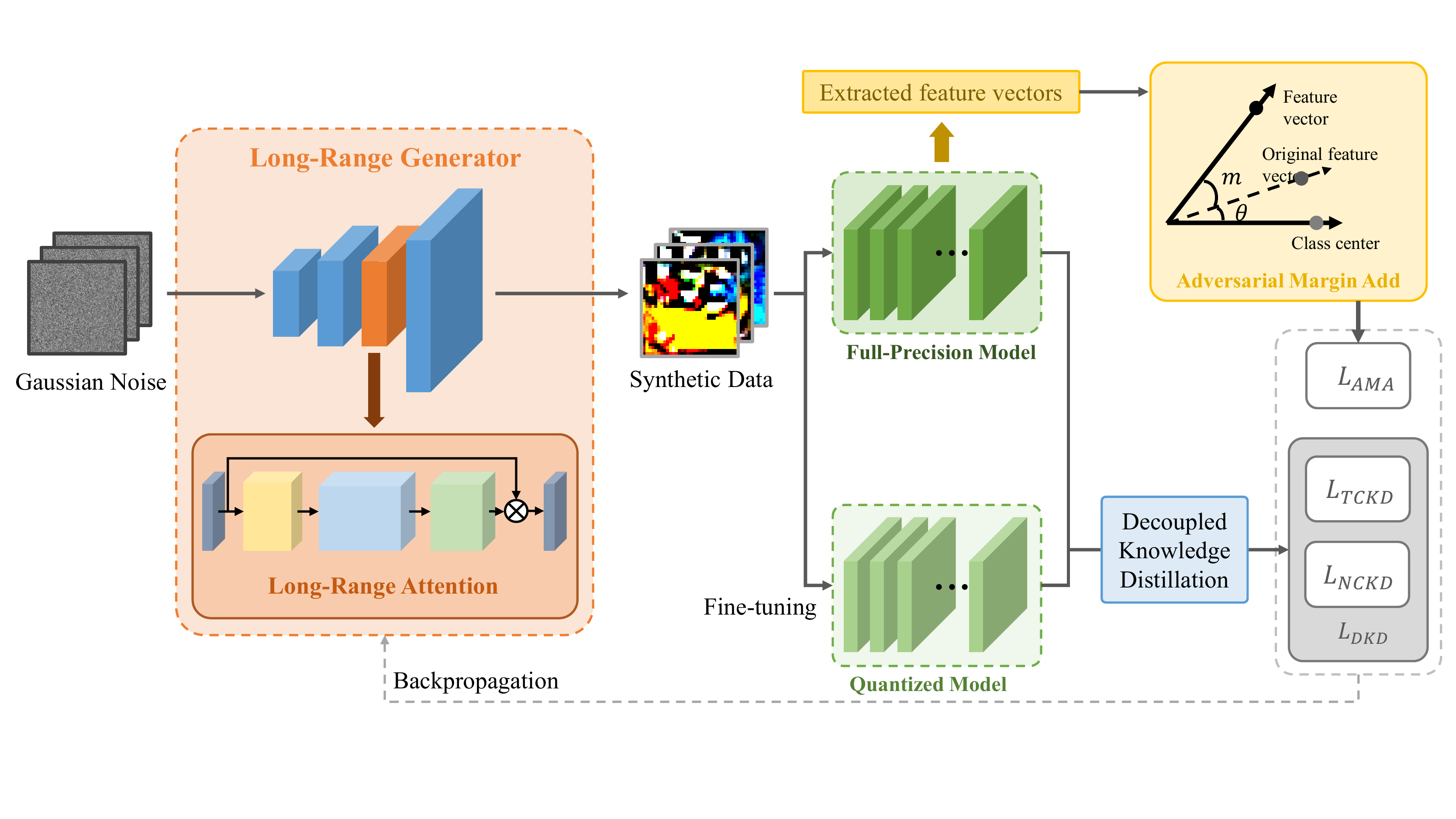}
    \vspace{-12mm}
   \caption{The pipeline of our LRQ, which includes three components, i.e., Long-Range Generator (LRG), Adversarial Margin Add (AMA), and Decoupled Knowledge Distillation (DKD). LRG aims at learning long-range information during synthetic data generation. AMA focus on enlarging intra-class heterogeneity via intra-class angular enlargement. DKD transfers knowledge from full-precision model.}
    \label{fig:main}
\end{figure*}
  \vspace{-3mm}
\section{Introduction}
With the fast development of technologies and applications of deep neural networks (DNNs)\cite{ji2021cnn,zhang2021data,wei2021deraincyclegan}, the need for computing power and memory keeps increasing. This therefore results in a challenging issue on how to deploy large DNN models into edge devices with limited computation resources. As a result, model compression methods came into being, such as model pruning \cite{yu2022width}, knowledge distillation \cite{he2022knowledge}, tensor decomposition \cite{rigamonti2013learning,gao2021dictionary} and network quantization, among which quantization is more easily supported by hardware \cite{jouppi2017datacenter}. Quantization can be further divided into two categories: Quantization Aware Training (QAT) and Post Training Quantization (PTQ). Specifically, QAT fine-tunes the weights of model and can maintain or even exceed the accuracy of the original full-precision model. In contrast, QAT needs a lot of original data and is more computationally expensive. And more the point, PTQ enables fast inference with small amount of data and does not require heavy computations. 

In reality, the access to the original data may be severely restricted for the considerations of data security and privacy, e.g., medical treatments, chat logs and confidential transactions. This poses great challenges for PTQ, although it only needs a small amount of original data. To address this issue, Zero-Shot Quantization (ZSQ) is therefore proposed, which can quantize a deep model without any original data \cite{zhang2021diversifying, choi2020data, cai2020zeroq, nagel2019data}. It is noteworthy that a large performance gap still exists between both ZSQ and PTQ, due to the absence of original data. Recently, some generative methods have been proposed to address this issue \cite{zhong2022intraq, xu2020generative, liu2021zero, he2021generative, choi2022s, gao2022clusterq}. By utilizing more information of the pretrained full-precision model, the generated synthetic data have similar distribution as original data, which lead to higher accuracy. For example, IntraQ \cite{zhong2022intraq} enhances the diversity of the generated data via local object reinforcement, which randomly crops a patch from input. Simultaneously, IntraQ retains intra-class heterogeneity via marginal distance constraint, which expects to form heterogeneous intra-class features. However, we think the performances of these methods are still limited and such performance gap roots in two main limitations:

\textbf{(1) Generator lacks long-range information to enhance data diversity.} 
In ZSQ, the diversity of the synthetic data determines the quantized model performance \cite{zhang2021diversifying}. To increase the diversity, data distribution needs to contain more comprehensive information. However, the generator used in current ZSQ methods cannot well handle long-range dependency due to the narrow receptive field of convolution kernel, which will cause information loss. Lacking long-range information prevents the synthetic data to accurately approximate the real data due to inadequate global features.

\textbf{(2) Weak intra-class heterogeneity within synthetic data.} Images contain different content information, even if they are picked from the same class. As a result, features of real data from the same class will also vary a lot. Fine-tuned with the synthetic data which are mostly homogeneous within the same class, the quantized model will have the limited generalization ability to the real-world data. Although IntraQ introduced a marginal distance constraint as a supervisory signal to guide the feature learning, we find that the intra-class distance are too small when encountering larger datasets and the loss of the marginal distance constraint always falls to zero during training.

In this paper, we therefore present a new zero-shot quantizer that sufficiently discovers the long-range information and enhance the intra-class heterogeneity during data generation, in order to improve the quantized model performance. The main contributions are summarized as follow: 

\begin{itemize}
\vspace{-2mm}
\item \textbf{Long-Range Zero-Shot Generative Deep Network Quantization (LRQ).} Technically, we introduce a new	Long-Range Generator (LRG) to learn the long-range information during synthetic data generation, and propose an Adversarial Margin Add (AMA) module to enhance intra-class heterogeneity. A decoupled knowledge distillation is also used to transfer knowledge between models. To the best of our knowledge, this is the first work to learn long-range information for enhancing the diversity of synthetic data in ZSQ. 
\vspace{-2mm}
\item \textbf{Long-Range Generator.} The generators used in current ZSQ methods often ignore the long-range dependency, so we introduce a long-range attention (LRA), which expands the originally small receptive field. A new long-range generator (LRG) is proposed to discover long-range dependencies and local contextual information, leading to higher diversity of synthetic data. As shown in Figure \ref{fig:attention}, our quantized model fine-tuned with the generated data has a strong ability to acquire information. Even with occlusion, it still can pay attention to the whole object instead of parts.
\vspace{-1mm}
\item \textbf{Adversarial Margin Add (AMA) Module.} To retain and enhance the intra-class heterogeneity, AMA module forces the intra-class angular enlargement between the feature vector and class center, which enlarges the intra-class distance. At the same time, such an operation increases the convergence difficulty of the loss function, which therefore forms an adversarial process. Effective discrepancy estimation is achieved due to this adversarial learning manner. Besides, a decoupled knowledge distillation (DKD) \cite{zhao2022decoupled} is also integrated to enhance the adversarial process, and help improve the quantized model performance.
\vspace{-1mm}
\item \textbf{SOTA Performance.} Based on discovering the long-range information and enhancing the intra-class heterogeneity, superior performance are obtained by our LRQ. For example, our LRQ achieves 56.87$\%$ top-1 accuracy on ImageNet when quantizing MobileNetV1 to 4-bit, which leads to an increase of 5.51$\%$ compared to the advanced IntraQ \cite{zhong2022intraq}. 

\end{itemize}

\section{Related Work}
In this section, we briefly review the PTQ and ZSQ methods which are closely-related to our model.
\subsection{Post-training Quantization (PTQ)}
PTQ methods aim to alleviate the performance deterioration from two aspects: 1) designing more sophisticated quantization methods; 2) introducing new loss functions for rounding optimization. Firstly, Liu et al. \cite{liu2021post} reduce the gap between full-precision weight vector and its low-bit version using a linear combination of multiple low-bit vectors. Analytical Clipping for Integer Quantization (ACIQ) with per-channel bit allocation is proposed in \cite{banner2019post}, which can not only obtain the clipping threshold, but also set different bit-widths for different channels to quantize. Secondly, instead of rounding the round-to-nearest when quantizing each weight in the convolution, AdaRound \cite{nagel2020up} adaptively decides whether to shift the floating-point value to the nearest right or left fixed-point value. BRECQ \cite{li2021brecq} proposes to use a block-wise reconstruction with Fisher information matrix based on the second-order analysis. QDrop incorporates activation to optimize weight rounding process. 

\subsection{Zero-Shot Quantization (ZSQ)}
Compared to PTQ, ZSQ methods can quantize model without real training data. DFQ \cite{nagel2019data} is the first to propose the idea of the ZSQ, which equalizes the range of data for each channel of two adjacent layer weights by the mathematical properties of activation function. One approach is to synthesize data by optimizing Gaussian noise, such as ZeroQ \cite{cai2020zeroq} that proves the batch normalization (BN) layer contains the statistical information of training set, and can generate the distilled data through matching the batch normalization statistics (BNS) information (e.g., mean and standard deviation) of all layers. 
Another approach is to synthesize data by generator. For example, GDFQ \cite{xu2020generative} first uses a Generator to generate data and train network jointly by the BNS loss and cross-entropy loss; AutoReCon \cite{zhu2021autorecon} introduces a reconstruction method of neural architecture search into the design of generator; ZAQ \cite{liu2021zero} introduces the adversarial learning into ZSQ through discrepancy estimation and knowledge transfer stages; AIT \cite{choi2022s} includes the CE loss and KL loss into the GDFQ training, where Gradient inundation (GI) is used to overcome the tricky issue of varying the parameter magnitudes encountered in the KL-only training. 
It is noted that the synthetic data constrained by the BN statistic have homogeneity issue at the distribution of classification results, which also has direct impact on the performance. To address this issue, DSG \cite{zhang2021diversifying} adds the relaxation constant to the original BNS loss function to solve the distribution homogeneity issue; Qimera \cite{choi2021qimera} enables the generator to synthesize boundary supporting samples by using superposed latent embeddings, thereby enhancing the quantization effect; IntraQ \cite{zhong2022intraq} enhances the diversity of the generated data by discovering intra-class heterogeneity; ClusterQ \cite{gao2022clusterq} solves this issue by clustering and aligning the feature distribution statistics to imitate the distribution of real data. Although these methods have attempted to solve the intra-class or inter-class homogeneity, the enhancement quality is still limited. Meanwhile, the problem of lacking long-range information is not taken into consideration.

\section{Methodology}
In this section, we introduce the framework of LRQ (see Figure
\ref{fig:main}), which aims at getting higher diversity of synthetic data and discovering long-range information. In general, our LRQ includes three main components, i.e., Long-Range Generator (LRG), Adversarial Margin Add (AMA), and Decoupled Knowledge Distillation (DKD).

\subsection{Preliminaries}
For better implementation on edge devices, we use an asymmetric uniform quantizer for network quantization. Given a floating-point value $x$ , the process of quantizing it to an $N$-bit integer $x_q$ can be expressed as:
\begin{equation}
{x_q} = round(\frac{x}{S} - Z), \ S = \frac{{\beta - \alpha}}{{{2^{N}} - 1}},
\end{equation}
where $S$ is the scaling factor for uniform mapping, $Z$ is the integer offset, $N$ is the bit width of  the quantized fixed-point value $x_q$, $\alpha$ and $\beta$ denote the lower and upper bounds of the clipping range. $round(\cdot)$ is the rounding operation which rounds mapped value to the nearest $N$-bit integer. Then, the dequantized value $\bar{x}$ is obtained by $\bar{x} = {x_q} \cdot S$. Due to the poor representation ability of low-bit value, quantization process introduces noise, requiring fine-tuning to recover the degraded performance.

Recently, data synthesis has attracted much attention for ZSQ with fine-tuning. Aiming at imitating the real-data distribution, majority of current methods generate fake images by making full use of the pretrained full-precision model $M_{FP}$. Technically, BNS loss $L_{BNS}$ is employed to constrain the activation distribution \cite{zhang2021diversifying, zhong2022intraq, gao2022clusterq}:
\begin{equation}
{L_{BNS}} = \sum\limits_{i = 1}^N {||\mu _S^k - \mu _P^k|| + ||\sigma _S^k - \sigma _P^k||},
\end{equation}
where $\mu _P^k$ and $\sigma _P^k$ are the mean and variance stored in the
$k$-th BN layer of the pre-trained full-precision model $M_{FP}$. Meanwhile, $\mu _S^k$ and $\sigma _S^k$ denote the running mean and variance in the $k$-th BN layer.

\subsection{Long-Range Generator}
Due to the restricted access to any real data in the setting of ZSQ, the quantized model performance mainly depends on the diversity of the synthetic data produced by the generator \cite{zhang2021diversifying}. This necessitates higher diversity within the synthetic data. Ordinary generators with small convolution kernel size are employed in previous generative ZSQ works including IntraQ \cite{zhong2022intraq}, which learn local information to generate fake data, but ignoring the long-range dependencies. This shortage restricts the diversity of synthetic data generation, eventually leads to the limited recovery of damaged information of the quantized model. As shown in Figure \ref{fig:attention}, we see that the model quantized by IntraQ represents a scattered and inaccurate attention to objects, especially when there exists occlusion. To better utilize the comprehensive knowledge from the pretrained model for data generation, long-range information learning should be considered into the generator design. Therefore, we propose a long-range generator LRG with attention mechanism.

\begin{figure}[t]
  \centering
   \includegraphics[width=\linewidth]{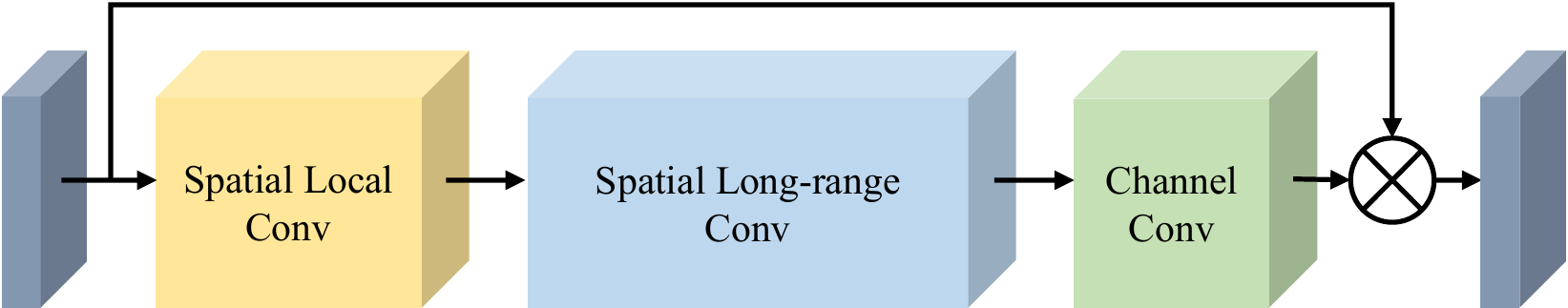}
   \caption{The process of long-range attention consists of three components: spatial local convolution, spatial long-range convolution, and channel convolution.}
   \label{fig:LRA}
\end{figure}

\begin{figure}[t]
  \centering
   \includegraphics[width=\linewidth]{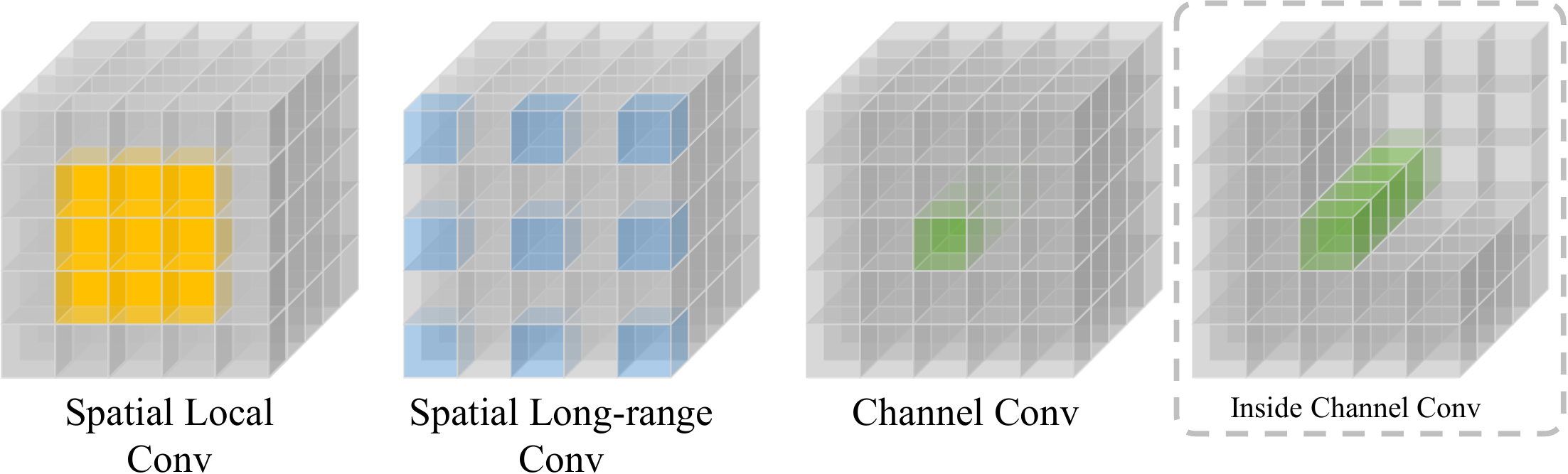}
   \vspace{-5mm}
   \caption{Illustration of the three components in the long-range attention. Each small cube indicates pixel in the layer input and colored cubes denote the pixels covered by the kernel. Noted that zero paddings are not presented for simplicity. Inside view is provided for the channel convolution. }
   \label{fig:kernel_diagram}
\end{figure}

The attention mechanism can concentrate the attention of a model to obtain features from most key parts of an image to obtain critical information, which has been widely used in image classification and other vision tasks \cite{guo2022attention}. To process the long-range dependencies with attention mechanism, the intuitive idea is to directly use large-kernel convolution which has a large receptive field. However, this manner leads to expensive computation cost. Inspired by the MobileNet \cite{howard2017mobilenets,sandler2018mobilenetv2} and Visual Attention Network (VAN) \cite{guo2022visual}, we use a long-range attention operation in our generator to learn long-range relationship. Different from the methods with self-attention \cite{liu2021swin,dosovitskiy2021an,jiang2021transgan,zhang2019self}, the design of our approach brings a more efficient training process and lower computational consumption. As shown in Figure \ref{fig:kernel_diagram}, to approximate a large convolution kernel of $K\times K$, we define dilation $d$ of 3 to control the size of receptive field, and the long-range attention consists of three components:

\begin{itemize}
    \vspace{-2mm}
    \item \textbf{Spatial Local Convolution.} Local spatial information can be well learned through $(2d - 1) \times (2d - 1)$ depth-wise convolution that is employed in the Spatial Local Convolution ($SL_{Conv}$), so that the texture of generated image can be effectively refined.
    \vspace{-2mm}
    
    \item \textbf{Spatial Long-range Convolution.} With the design of $\left\lceil {K/d} \right\rceil \times \left\lceil {K/d} \right\rceil$ depth-wise dilated convolution, the receptive field of the long-range attention can be expanded to process long-range dependencies better. The spatial long-range convolution ($SLR_{Conv}$) contributes the attention on those long-range pixels whereas missing the information of neighbor pixels which can be compensated by the spatial local convolution.
    \vspace{-2mm}
    
    \item \textbf{Channel Convolution.} In this design, $1\times1$ convolution is used in the channel convolution ($C_{Conv}$) to reduce the channel of the feature maps without severe information damage and thus leading to more efficient training process. In addition, channel convolution allows cross-channel information learning which facilitates the diversity enhancement of features. 
    \vspace{-2mm}
\end{itemize}
\begin{figure*}[t]
	\centering
	\begin{subfigure}[b]{0.245\linewidth}
		\includegraphics[width=1\linewidth]{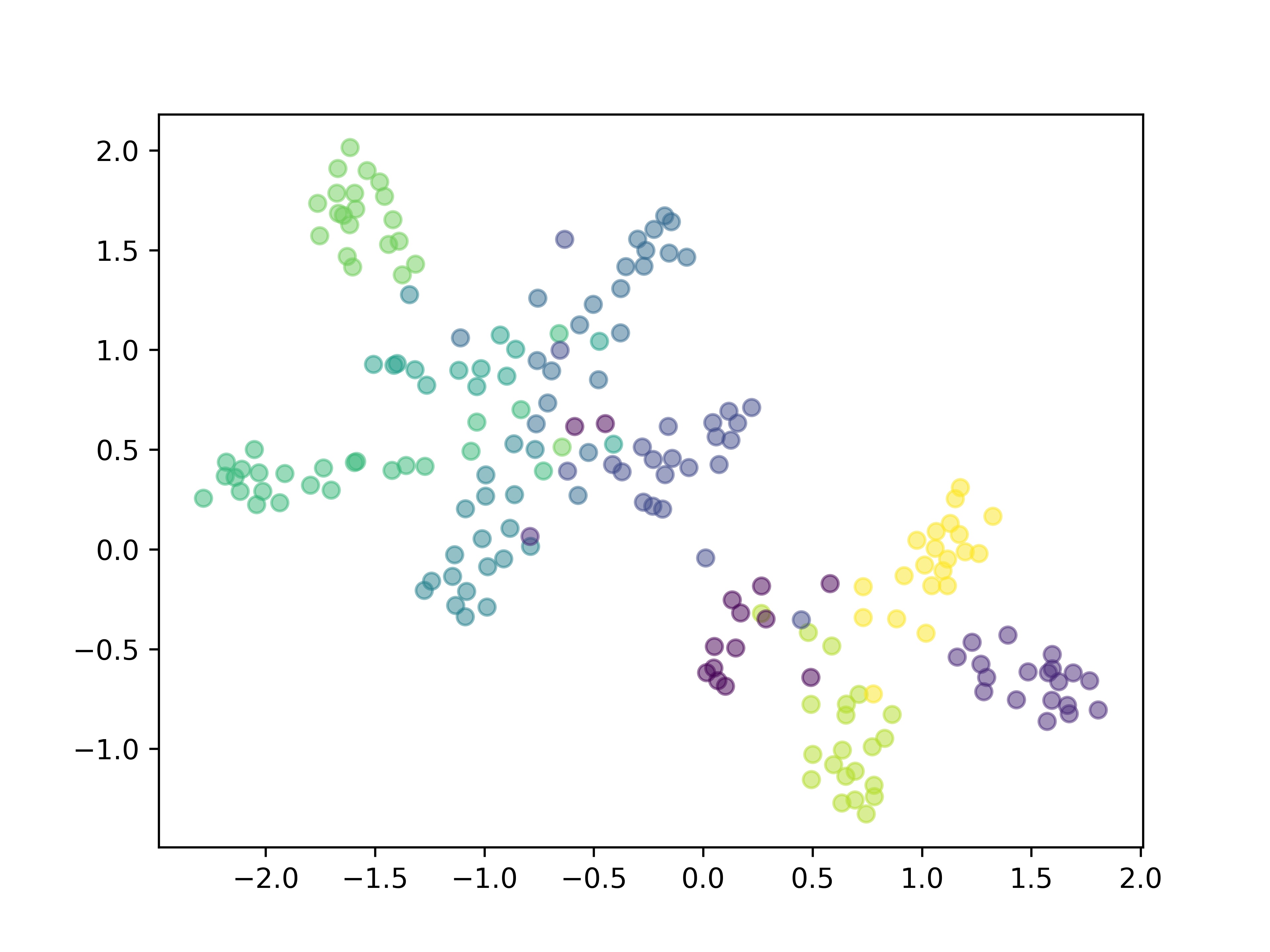}
		\caption{}
	\end{subfigure}
	\begin{subfigure}[b]{0.245\linewidth}
		\includegraphics[width=1\linewidth]{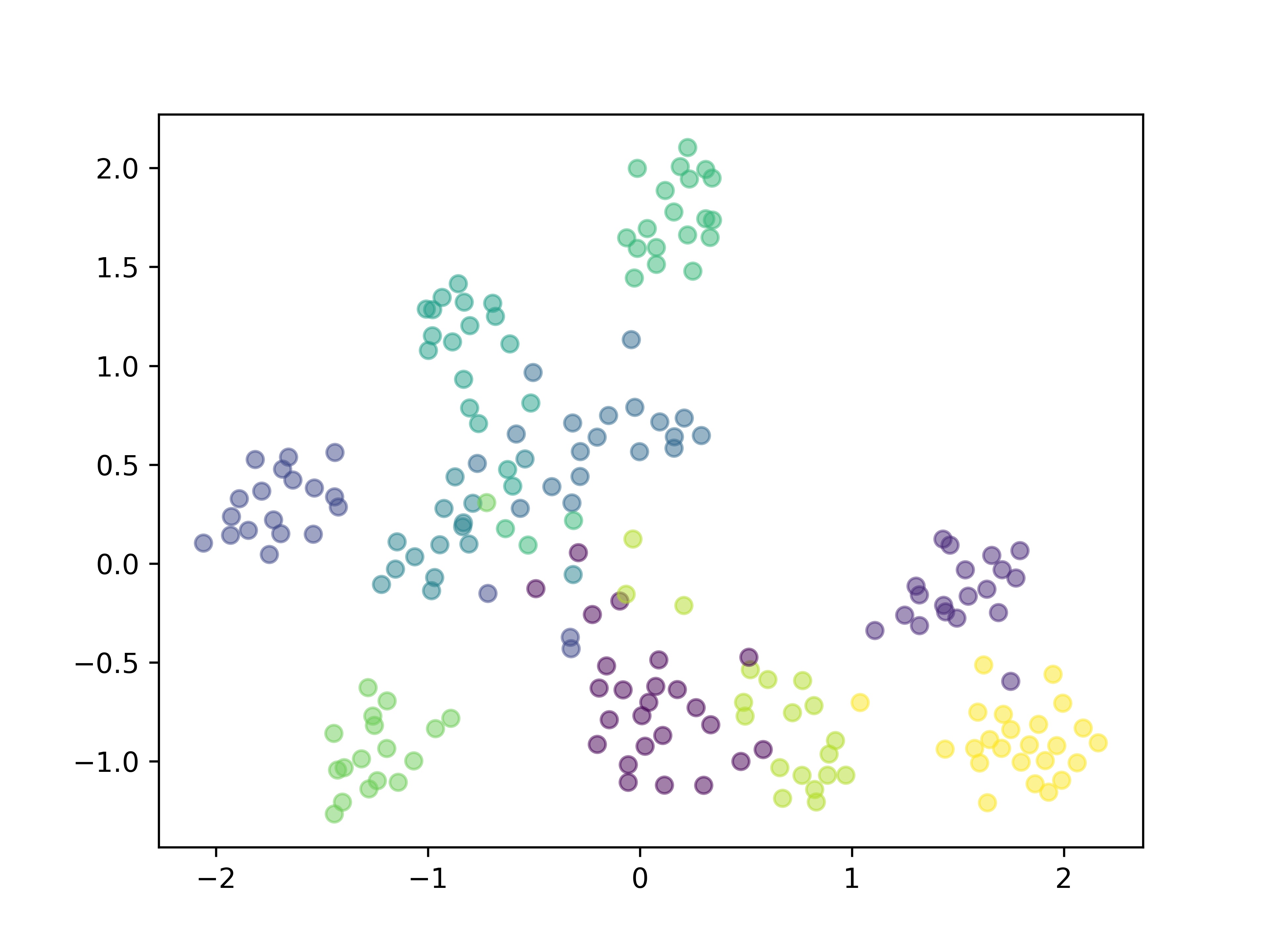}
		\caption{}
	\end{subfigure}
	\begin{subfigure}[b]{0.245\linewidth}
		\includegraphics[width=1\linewidth]{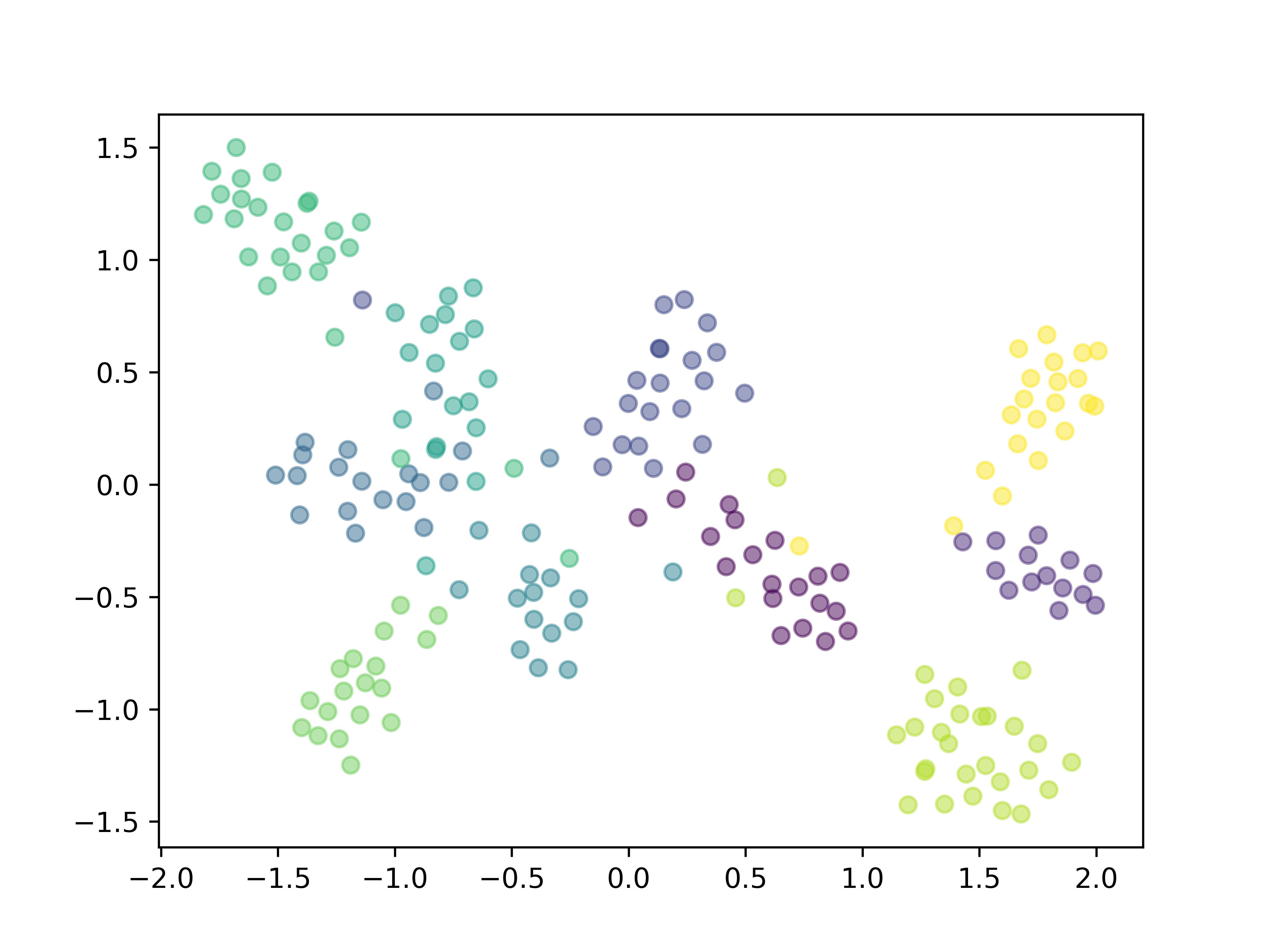}
		\caption{}
	\end{subfigure}
	\begin{subfigure}[b]{0.245\linewidth}
		\includegraphics[width=1\linewidth]{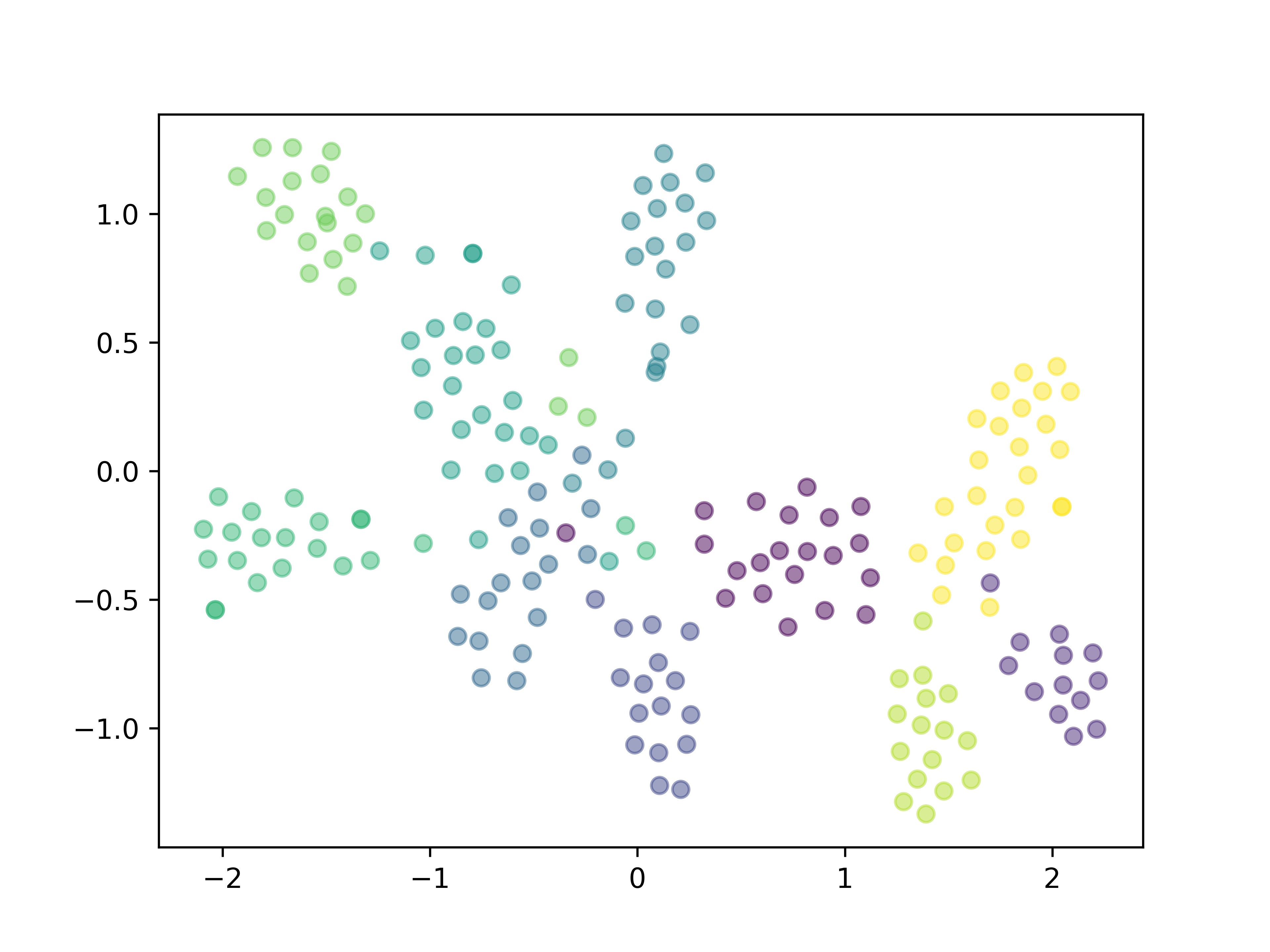}
		\caption{}
	\end{subfigure}
 \vspace{-2mm}
	\caption{t-SNE visualization results of the last layer features in ResNet-20 model getting on CIFAR-10. From (a) to (d) correspond to the features of GDFQ \cite{xu2020generative}, ClusterQ\cite{gao2022clusterq}, IntraQ\cite{zhong2022intraq} and our LRQ. Clearly, our method obtains higher inter-class sepration than other three.}
	\label{fig:tsne}
	 \vspace{-1mm}
\end{figure*}

Under this design, the long-range dependencies can be well handled with much less calculation. The diagram of the long-range attention is shown in Figure \ref{fig:kernel_diagram}. After obtaining the long-range relationship, we can estimate the importance of a point and generate attention map. As shown in Figure \ref{fig:LRA}, the long-range attention can be defined as
\begin{equation}
LA = C_{Conv}(SLR_{Conv}(SL_{Conv}(V))),
\end{equation}
where $V$ denotes the input feature vector, $SL_{Conv}$ denotes the spatial local convolution, $SLR_{Conv}$ denotes the spatial long-range convolution with dilation $d$, $C_{Conv}$ is the channel convolution, and $LA$ denotes long-range attention map. Then, the output features can be computed as
\begin{equation}
\tilde{V} = Attention \otimes V,
\end{equation}
where $\tilde{V}$ is output features and $\otimes$ is element-wise product. 

Compared to the methods with self-attention, our LRG with long-range attention not only learns global dependencies but also local contextual information, which are incredibly important to generate diverse samples with rich and detailed textures, which will be closer to the real data. Moreover, as a fixed-shape kernel, the long-range attention leads to easier training of the generator with less synthetic data.

\subsection{Adversarial Margin Add (AMA)}
Suppose the generated synthetic data have higher diversity, the pre-trained full-precision network $M_{FP}$ can obtain more feature information during training. In contrast, if the model is fine-tuned with the synthetic data which are mostly homogeneous within each class, the quantized model will have low accuracy and weak generalization ability to real-world data. To classify all the classes more accurately, we expect $M_{FP}$ to form class-related features with enhanced intra-class discrimination. Therefore, AMA module incorporates a new margin into the calculation of cosine similarity between feature vector and class center. AMA increases the intra-class distance by enlarging the angle. This margin increases the convergence difficulty of the loss function, and further increases the training difficulty of the model. However, the goal of loss function is to reduce the difficulty to convergence. That is, the training of AMA is opposite to that of the original loss function, forming an adversarial process. Note that this adversarial process makes training more difficult, but it can benefit the performance.

Let $F(\bar I_i)$ denote the feature vector of $\bar I_i$ produced by the pre-trained full-precision network $M_{FP}$, and $C(\bar{I})$ denote the class center of $\bar{I}$. Supposing that the label of $\bar{I}$ is $c$ and $M_c$ is a collection of synthetic data belonging to the class $c$, we can define the class center as the mean feature vector of all synthetic images in $M_c$:
\begin{equation}
C(\bar I) = \frac{1}{{{M_c}}}\sum\limits_{i = 1}^{|{M_c}|} {F({{\bar I}_i})} , {\bar I_i} \in {M_c}.
\end{equation}

As the embedding features are distributed around each feature center on the hypersphere, we add an additive angular margin penalty $m$ between $F(\bar{I})$ and $C(\bar{I})$ to enhance the intra-class compactness:
\begin{equation}
\begin{aligned}
{L_{AMA}}(\theta ) &= \max ({\lambda _l} - \cos (\theta  + m),0) \\
                    & + \max(\cos(\theta+m)-{\lambda_u},0),
\end{aligned}
\end{equation}
where $\cos(\theta)=\cos{(F(\bar{I}),C(\bar{I}))}$ returns the cosine distance, $m$ denotes the margin we added, $\lambda _l$ and $\lambda _u$ are the lower and upper bounds of the cosine distance between $\bar{I}$ and its class center. Specifically, the margin $m$ is forced into the calculation of cosine similarity. To highlight our motivation on the intra-class heterogeneity of semantic features, we conduct some visualization experiments on the DNN features to observe the dynamic transformation of this heterogeneity over different models, as illustrated in Figure \ref{fig:tsne}. Compared to the classification results of other two on the CIFAR-10 dataset, our LRQ obtains greater intra-class distances than them.

\subsection{Decoupled Knowledge Distillation}
To well handle the adversarial process in AMA module, the training of the quantized model should be enhanced. Previous works have used traditional knowledge distillation (KD) \cite{xu2020generative,gao2022clusterq,zhong2022intraq}, hoping to transfer knowledge from full precision model to the quantized models. However, with the improvement of quantization performance, this method is difficult to enhance the quantization effect, but instead occupies part of the computing power. Therefore, we use a more effective  method called Decoupled Knowledge Distillation (DKD) \cite{zhao2022decoupled}.
Traditional KD uses the KL-Divergence as loss function, while DKD reformulates the classical KD with binary probabilities and the probabilities among non-target classes. The loss function of DKD is defined as
\begin{equation}
{L_{dkd}} = \alpha {L_{TCKD}} + \beta {L_{NCKD}},
\end{equation}
where $L_{TCKD}$ (Target Class Knowledge Distillation) represents the similarity between the teacher’s and student’s binary probabilities of the target class. Meanwhile, $L_{NCKD}$ (Non-Target Class Knowledge Distillation) represents the similarity between the teacher’s and student’s probabilities among non-target classes. $\alpha$ and $\beta$ are the weights to balance the importance of $L_{TCKD}$ and $L_{NCKD}$, respectively. Through a more effective training method, DKD can better transfer the knowledge learned by the previous AMA module to the quantized model. This further enables the quantized model to learn knowledge that preserves intra-class heterogeneity. Following \cite{zhao2022decoupled}, $\alpha$ is set to 1 and $\beta$ is set to 8.

\subsection{Training Process}
The training of LRQ includes data generation for synthetic data and network fine-tuning for quantized model.
\subsubsection{Data Generation}
A standard Gaussian distribution $I$ is used as input data. This data generation process transfers $I$ to $\bar{I}$ via Long-Range Generator to well match the distribution of real data, in particular with the long-range information. To this end, as shown in Figure \ref{fig:main}, we first use the long-range generator to derive $\bar{I}$. Next, the loss of batch normalization statistics (BNS) and our AMA module by $\bar{I}$ are computed as
\begin{equation}
L = L_{BNS} + L_{AMA}.
\end{equation}

\subsubsection{Network Fine-Tuning}
We fine-tune the quantized model with the cross-entropy loss $CE(\cdot)$ using the generated fake images $\bar{I}$:
\begin{equation}
L_{CE} = CE(Q(\bar{I}),y),
\end{equation}
where $y$ is the true label of real data. We transfer the output of the pre-trained $M_{FP}$ to $M_Q$ using $L_{dkd}$ as follows:
\begin{equation}
L_{DKD} = L_{dkd}(Q(\bar{I}),P(\bar{I})).
\end{equation}

Combining them forms the overall loss function during the process of fine-tuning $M_Q$: 
\begin{equation}
L = L_{CE} + \lambda{L_{DKD}},
\end{equation}
where $\lambda$ balances the importance of $L_{CE}$ and $L_{DKD}$.

\section{Experiments}
In this section, we evaluate our proposed LRQ on several image datasets, along with illustrating the comparison results to several closely-related ZSQ methods.

\subsection{Implementation Details}
The experiments are conducted on the validation of ImageNet \cite{russakovsky2015imagenet} and CIFAR-10/100 \cite{krizhevsky2009learning}, which are implemented with Pytorch \cite{paszke2019pytorch}. We quantized the ResNet-18 \cite{he2016deep}, MobileNetV1 \cite{howard2017mobilenets} and MobileNetV2 \cite{sandler2018mobilenetv2} over ImageNet, and ResNet-20 \cite{he2016deep} for CIFAR-10/100. Adam \cite{kingma2014adam} is applied during data generation, while the momentum and the initial learning rate are set to 0.9 and 0.5. The learning rate is decayed by 0.1 for every 1,000 updates to the synthetic data. And the batch size is set to 256. There exist three hyperparameters during the process of data generation, including $K$, $m$ and $\lambda$, which are respectively set to 21, 0.6 and 0.9. 

The synthetic images produced by LRG to quantize the model using the stochastic gradient descent (SGD) with Nesterov \cite{1983A}. 150 fine-tuning epochs are provided, and the weight decay is set to $10^{-4}$. For CIFAR-10/100 and ImageNet, the batch size for fine-tuning is 256 and 16, respectively. In addition, the learning rate is $10^{-4}$ and $10^{-6}$ for CIFAR-10/100 and ImageNet respectively. And it decays by 0.1 every 100 epochs.

\subsection{Performance Comparison}
\subsubsection{Results on ImageNet}
We first analyze the performance of several ZSQ methods on the ImageNet dataset, including GZNQ \cite{he2021generative}, ClusterQ \cite{gao2022clusterq}, Qimera \cite{choi2021qimera}, IntraQ \cite{zhong2022intraq} and our LRQ. MobileNetV1, MobileNetV2, and ResNet-18 are considered as quantized networks. We display the 5-bit and 4-bit results to show the effectiveness of our LRQ.

\textbf{Resnet-18.} The results of experiments on ResNet-18 are provided in Table \ref{tab:resnet18}, where W$m$A$n$ indicates that the weights and activations are quantized to $m$-bit and $n$-bit respectively, and +IL stands for adding the inception loss for DSQ and ZeroQ. In the case of 5-bit, our LRQ slightly outperforms the current SOTA method IntraQ. For 4-bit quantization, a noticeable increase is obtained by our LRQ over other competitors, followed by IntraQ, and both are significantly superior to the remaining methods.  

\textbf{MobileNetV1/V2.} Tables \ref{tab:mobileV1} and \ref{tab:mobileV2} describes the experimental results of each method on MobileNetV1/V2. We see that our proposed LRQ obtains the best performance in quantizing MobileNetV1 and MobileNetV2. The performance increase is most noticeable in 4-bit quantization scenarios. For instance, our LRQ obtains 5.48\% accuracy improvement over the advanced IntraQ when MobileNetV1 are quantized into 4-bit.

\setlength\tabcolsep{17pt}
\begin{table}[t]
    \renewcommand{\arraystretch}{1.08}
	\centering
	\caption{Results of ResNet-18 on ImageNet.} 
	\vspace{-2mm}
	\begin{tabular}{ccc}
		\hline
		Bit-width & Method & Acc. (\%) \\
		\hline
		W32A32 & Full-precision & 71.47 \\
		\hline
		\multirow{8}*{W5A5}
        & Real data & 70.31\\
        & GDFQ      & 66.82\\
        & DSG       & 69.53\\
        & ZeroQ     & 69.65\\
        & DSG+IL    & 69.53\\
        & ZeroQ+IL  & 69.72\\
        & IntraQ    & 69.94\\
        & LRQ (Ours) & \textbf{70.07}\\
        \hline
		\multirow{12}*{W4A4}
        & Real data & 67.89\\
        & GDFQ      & 60.60\\
        & DSG       & 60.12\\
        & ZeroQ     & 60.68\\
        & AutoReCon & 61.60\\
        & ZeroQ+IL  & 63.38\\
        & DSG+IL    & 63.11\\
        & Qimera    & 63.84\\
        & ClusterQ  & 64.39\\
        & GZNQ      & 64.50\\
        & IntraQ    & 66.47\\
        & LRQ (Ours) & \textbf{67.19}\\
		\hline
	\end{tabular}
	\label{tab:resnet18}
		\vspace{-3mm}
\end{table}

\setlength\tabcolsep{17pt}
\begin{table}[t]
    \renewcommand{\arraystretch}{1.08}
	\centering
	\caption{Results of MobileNetV1 on ImageNet.} 
	\vspace{-2mm}
	\begin{tabular}{ccc}
		\hline
		Bit-width & Method & Acc. (\%) \\
		\hline
		W32A32 & Full-precision & 73.39 \\
		\hline
		\multirow{8}*{W5A5}
        & Real data & 69.87\\
        & GDFQ      & 59.76\\
        & ZeroQ     & 61.95\\
        & DSG       & 64.18\\
        & DSG+IL    & 66.61\\
        & ZeroQ+IL  & 67.11\\
        & IntraQ    & 68.17\\
        & LRQ (Ours) & \textbf{68.60}\\
        \hline
		\multirow{8}*{W4A4}
        & Real data & 59.66\\
        & ZeroQ     & 20.96\\
        & DSG       & 21.14\\
        & ZeroQ+IL  & 25.43\\
        & GDFQ      & 28.64\\
        & DSG+IL    & 42.19\\
        & IntraQ    & 51.36\\
        & LRQ (Ours) & \textbf{56.87}\\
		\hline
	\end{tabular}
	\label{tab:mobileV1}
\end{table}

\setlength\tabcolsep{17pt}
\begin{table}[t]
    \renewcommand{\arraystretch}{1.08}
	\centering
	\caption{Results of MobileNetV2 on ImageNet.} 
	\vspace{-2mm}
	\begin{tabular}{ccc}
		\hline
		Bit-width & Method & Acc. (\%) \\
		\hline
		W32A32 & Full-precision & 73.03 \\
		\hline
		\multirow{8}*{W5A5}
        & Real data & 72.01\\
        & GDFQ      & 68.14\\
        & DSG       & 70.85\\
        & ZeroQ     & 70.88\\
        & DSG+IL    & 70.87\\
        & ZeroQ+IL  & 70.95\\
        & IntraQ    & 71.28\\
        & LRQ (Ours) & \textbf{71.37}\\
        \hline
		\multirow{10}*{W4A4}
        & Real data & 67.90\\
        & GDFQ      & 51.30\\
        & DSG+G     & 54.66\\
        & GZNQ      & 53.53\\
        & ZeroQ     & 59.39\\
        & DSG       & 59.04\\
        & ZeroQ+IL  & 60.15\\
        & DSG+IL    & 60.45\\
        & IntraQ    & 65.10\\
        & LRQ (Ours) & \textbf{65.62}\\
		\hline
	\end{tabular}
	\label{tab:mobileV2}
\end{table}

\subsubsection{Results on CIFAR-10/100}
We also compare each method on the CIFAR-10/100 datasets. Because CIFAR-10 and CIFAR-100 are relatively small, the ultra-low precisions of 4-bit and 3-bit are utilized in quantizing ResNet-20. From the comparison results in Tables \ref{tab:resnet20_cifar10} and \ref{tab:resnet20_cifar100}, our LRQ achieves better performance compared to other related methods on both CIFAR-10 and CIFAR-100 datasets. Specifically, our LRQ improves the top-1 accuracy when quantizing the model to 3-bit on CIFAR-10 by 6.21\% in comparison to the IntraQ. Improvements are also obtained in the 4-bit case. 

\setlength\tabcolsep{17pt}
\begin{table}[t]
    \renewcommand{\arraystretch}{1.08}
	\centering
	\caption{Results of ResNet-20 on CIFAR-10.} 
	\vspace{-2mm}
	\begin{tabular}{ccc}
		\hline
		Bit-width & Method & Acc. (\%) \\
		\hline
		W32A32 & Full-precision & 94.03 \\
		\hline
		\multirow{9}*{W4A4}
        & Real data & 92.52\\
        & GDFQ      & 90.25\\
        & ZeroQ     & 84.68\\
        & DSG       & 88.74\\
        & DSG+IL    & 88.93\\
        & ZeroQ+IL  & 89.66\\
        & GZNQ      & 91.30\\
        & IntraQ    & 91.49\\
        & LRQ (Ours) & \textbf{91.55}\\
        \hline
		\multirow{8}*{W3A3}
        & Real data & 87.94\\
        & GDFQ      & 71.10\\
        & ZeroQ     & 29.32\\
        & DSG       & 32.90\\
        & DSG+IL    & 48.99\\
        & ZeroQ+IL  & 69.53\\
        & IntraQ    & 77.07\\
        & LRQ(Ours) & \textbf{83.28}\\
		\hline
	\end{tabular}
	\label{tab:resnet20_cifar10}
\end{table}

\begin{figure*}[t]
  \centering
   \includegraphics[width=\linewidth]{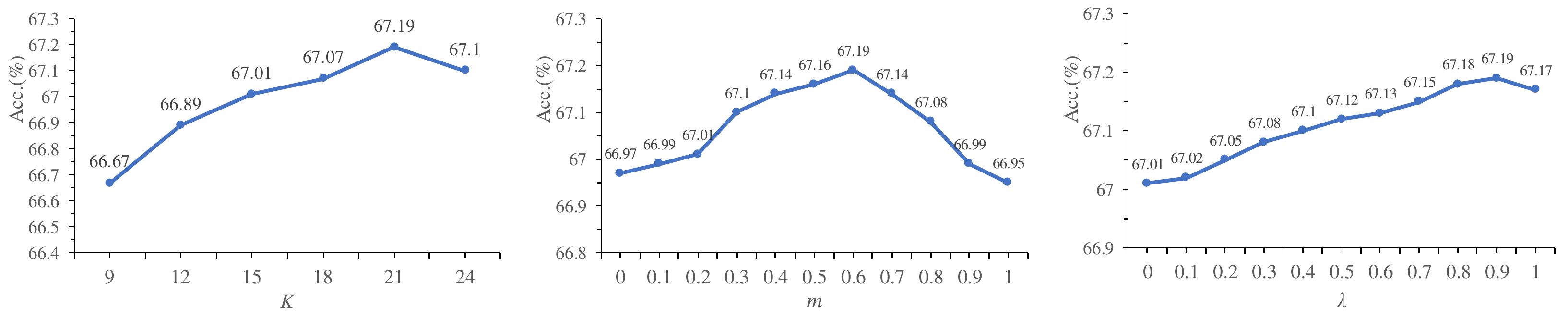}
   \vspace{-5mm}
   \caption{Ablation study on different hyper-parameters of our LRQ on ImageNet.}
   \label{fig:hyper}
\end{figure*}

\setlength\tabcolsep{17pt}
\begin{table}[t]
    \renewcommand{\arraystretch}{1.08}
	\centering
	\caption{Results of ResNet-20 on CIFAR-100.} 
	\vspace{-2mm}
	\begin{tabular}{ccc}
		\hline
		Bit-width & Method & Acc. (\%) \\
		\hline
		W32A32 & Full-precision & 70.33 \\
		\hline
		\multirow{9}*{W4A4}
        & Real data & 66.80\\
        & GDFQ      & 63.58\\
        & DSG       & 62.36\\
        & ZeroQ     & 58.42\\
        & DSG+IL    & 62.62\\
        & ZeroQ+IL  & 63.97\\
        & GZNQ      & 64.37\\
        & IntraQ    & 64.98\\
        & LRQ(Ours) & \textbf{65.67}\\
        \hline
		\multirow{8}*{W3A3}
        & Real data & 56.26\\
        & GDFQ      & 43.87\\
        & DSG       & 25.48\\
        & ZeroQ     & 15.38\\
        & DSG+IL    & 43.42\\
        & ZeroQ+IL  & 26.35\\
        & IntraQ    & 48.25\\
        & LRQ(Ours) & \textbf{48.47}\\
        \hline
	\end{tabular}
	\label{tab:resnet20_cifar100}
\end{table}

\subsection{Ablation Study}
We carry out ablation studies of the different LRQ hyper-parameters and components in this part. On ImageNet, all layers of ResNet-18 are quantized to 4-bit for the top-1 accuracy experiments.

\textbf{Effect of hyper-parameters.} We explore the impact of hyper-parameters $K$, $m$ and $\lambda$. We fix two and tune the other, and the results are shown in Figure \ref{fig:hyper}. Finally, an optimal combination of parameters is determined for all experiments, i.e., $K=21$, $m=0.6$ and $\lambda=0.9$. 

\textbf{Effect of components.} We further investigate the impact of our LRG, AMA and DKD modules in Table \ref{tab:Components}. We see that when the modules are individually added to synthesize the fake images, accuracies are always increased. Specifically, LRG  improves the performance more significantly, comparded to AMA and DKD modules. This is reasonable, since long-range information is very important for the synthetic images. When AMA and DKD are utilized together, the performance keeps improving. The best result can be achieved when they are all applied.
\setlength\tabcolsep{17pt}
\begin{table}[t]
    \renewcommand{\arraystretch}{1.06}
	\centering
	\caption{Ablation study on different components of our LRQ.} 
	\vspace{-2mm}
	\begin{tabular}{cccc}
		\hline
		LRG & AMA & DKD & Acc.(\%) \\
		\hline
         \ding{56} & \ding{56} & \ding{56} & 66.14\\
         \ding{52} & \ding{56} & \ding{56} & 66.88\\
         \ding{56} & \ding{52} & \ding{56} & 66.36\\
         \ding{56} & \ding{56} & \ding{52} & 66.27\\
         \ding{52} & \ding{52} & \ding{56} & 67.09\\
         \ding{52} & \ding{56} & \ding{52} & 66.97\\
         \ding{56} & \ding{52} & \ding{52} & 66.43\\
         \ding{52} & \ding{52} & \ding{52} & 67.19\\
        \hline
	\end{tabular}
	\label{tab:Components}
\end{table}

\section{Conclusion}
We have discussed the issues of achieving higher diversity of synthetic data and enhancing intra-class heterogeneity for zero-shot quantization. We proposed a novel long-range zero-shot generative quantization method. The proposed method is designed to learn more long-range information, where a newly-designed long-range attention can offer more global features for enhancing the diversity of synthetic data. In order to increase the intra-class distance, a new adversarial margin add module is further designed to enlarge the intra-class angle. Extensive experiments show the superiority and effectiveness of our method. In the future, we will explore more effective ways to further improve the diversity of synthetic data and intra-class heterogeneity. Quantizing deep low-level vision model \cite{zhang2022deep,wei2021deraincyclegan} for deployment on edge devices is also an interesting future work. 

\newpage

\section*{Acknowledgment}
This work is partially supported by the National Natural Science Foundation of China (62072151, 61932009, 61822701, 62036010, 72004174), and the Anhui Provincial Natural Science Fund for Distinguished Young Scholars (2008085J30). Zhao Zhang is the corresponding author of this paper.

{\small
\bibliographystyle{ieee_fullname}
\balance
\bibliography{myegbib}
}

\end{document}